\documentclass[journal]{IEEEtran}
\usepackage{graphicx}
\usepackage{easyReview}
\usepackage{amsmath}
\usepackage{amsfonts}
\usepackage{algorithm}
\usepackage{algorithmic}
\usepackage{makecell}
\usepackage{booktabs}
\setcellgapes{3pt}
\usepackage{multirow}
\setcellgapes{3pt}

\usepackage[justification=centering]{caption}
\usepackage{color}
\usepackage{verbatim}
\usepackage{amssymb}

\usepackage{cite}

\ifCLASSINFOpdf
\else
\fi
\begin{document}
%
\title{Deep progressive reinforcement learning-based flexible resource scheduling framework for IRS and UAV-assisted MEC system}

\author{Li Dong, Feibo Jiang, Minjie Wang, Yubo Peng and Xiaolong Li
	\thanks{
			This work was supported in part by the National Natural Science Foundation
			of China under Grant nos. 41904127 and 41604117. This work was also supported in part by the Open Project of Xiangjiang Laboratory (No.22XJ03011) and the Scientific Research Fund of Hunan Provincial Education Department(No.22B0663). (Corresponding author: Feibo Jiang and Xiaolong Li.)

		 Li Dong (Dlj2017@hunnu.edu.cn) is with School of Computer Science, Hunan University of Technology and Business, Changsha, China, and also with Xiangjiang Laboratory, Changsha, China. 
		 
		 Feibo Jiang (jiangfb@hunnu.edu.cn) is with Hunan Provincial Key Laboratory of Intelligent Computing and Language Information Processing, Hunan Normal University, Changsha, China. 
		 
		 Minjie Wang (minjie@hunnu.edu.cn) is with College of Information Science and Engineering, Hunan Normal University, Changsha, China.
		 
		  Yubo Peng (pengyubo@hunnu.edu.cn) is with College of Information Science and Engineering, Hunan Normal University, Changsha, China.
		  
		  Xiaolong Li (xlli@guet.edu.cn) is with Xiangjiang Laboratory, Changsha, China, and also with School of Computer Science, Hunan University of Technology and Business, Changsha, China. 
	}
}

\markboth{Submitted for Review}%
{Shell \MakeLowercase{\textit{et al.}}: Bare Demo of IEEEtran.cls for IEEE Journals}
%



\maketitle

\begin{abstract}
The intelligent reflection surface (IRS) and unmanned aerial vehicle (UAV)-assisted mobile edge computing (MEC) system is widely used in temporary and emergency scenarios. Our goal is to minimize the energy consumption of the MEC system by jointly optimizing UAV locations, IRS phase shift, task offloading, and resource allocation with a variable number of UAVs. To this end, we propose a Flexible REsource Scheduling (FRES) framework by employing a novel deep progressive reinforcement learning which includes the following innovations: Firstly, a novel multi-task agent is presented to deal with the mixed integer nonlinear programming (MINLP) problem. The multi-task agent has two output heads designed for different tasks, in which a classified head is employed to make offloading decisions with integer variables while a fitting head is applied to solve resource allocation with continuous variables. Secondly, a progressive scheduler is introduced to adapt the agent to the varying number of UAVs by progressively adjusting a part of neurons in the agent. This structure can naturally accumulate experiences and be immune to catastrophic forgetting. Finally, a light taboo search (LTS) is introduced to enhance the global search of the FRES. The numerical results demonstrate the superiority of the FRES framework which can make real-time and optimal resource scheduling even in dynamic MEC systems.

\end{abstract}


%
\IEEEpeerreviewmaketitle

\section{Introduction}

In recent years, computation-intensive applications such as virtual reality (VR) and augmented reality (AR) have grown rapidly, which makes the user equipment (UE) difficult to process these tasks locally. Mobile edge computing (MEC) addresses this issue by relocating computing and communication services closer to users, enabling UEs to offload computation-intensive tasks to nearby MEC servers. This offloading alleviates the computational burden and reduces the energy consumption of UEs \cite{RN1}.

However, deploying fixed MEC servers in terrestrial areas may not be considered economically viable, especially in temporary situation\cite{jiang2020ai}. 
Unmanned aerial vehicles (UAVs) are designed to be highly maneuverable, allowing them to navigate through tight spaces, fly at low altitudes, and perform precise aerial movements  \cite{9736441}.
UAV-assisted MECs have been proposed to provide computation and communication services to ground UEs seamlessly and proficiently. Whereas the wireless channels between UAVs and terrestrial UEs may be blocked by trees or high buildings.
 
Recently, the intelligent reflecting surface (IRS)\cite{RN3} has been proposed and received considerable attention. The quality of signal transmission can be improved by laying IRS on the surface of high buildings. 
Since each component in IRS can reflect the transmission signal, the transmission data rate can be greatly improved. In addition, the reflective elements of the IRS are generally passive, and the reflected signal does not require any signal processing capacities, which is more energy-efficient than the traditional transmission technologies. 
In UAV-assisted MEC systems, the IRS can be attached on the buildings or on the high places to enhance the channel between the UAV and UEs when the connections are blocked by buildings in temporary and emergency scenarios.

Although IRS and UAV-assisted MEC is promising, some challenges should be addressed. First, with the increase of UEs, a single UAV may not be sufficient to support a large number of computing tasks, and multiple UAVs may require to be deployed. However, with the changing number of UAVs and UEs, the communication environment is expected to become increasingly complex. Second, in the wireless fading environment, the wireless channel conditions which may change with time, can affect the resource scheduling of MEC systems\cite{RN4}, especially considering the IRS in the MEC system. Third, the offloading decision is generally an integer variable, and the resource allocation is the continues variable. It is an overall mixed integer nonlinear programming (MINLP) problem, which is difficult to be addressed by traditional methods.  

To address the above challenges, in this paper, we consider the MEC system assisted by IRSs and UAVs in the dynamic environment. Our objective is to design a deep progressive reinforcement learning based Flexible REsource Scheduling (FRES) framework to minimize the energy consumption of the MEC system with a variable number of UAVs. This problem requires not only a transfer learning method without catastrophic forgetting, but also an efficient solver for the MINLP problem. Compared with the existing resource scheduling methods, we have the following contributions:
\begin{enumerate}	
\item \emph{System Model}: We design an IRS and UAV-assisted MEC system in dynamic environments, and formulate the optimization problem to minimize the energy consumption of all UEs and UAVs by optimizing the UAV location, phase-shifted diagonal matrix, offloading decision and resource allocation.

\item \emph{Multi-task agent}: We propose a FRES framework with deep reinforcement learning (DRL) for offloading decision and resource allocation. In the FRES framework, a novel multi-task agent is designed to address the MINLP problem. The multi-task agent has two type of output heads, in which a classified head is applied to make offloading decision with integer variables while a fitting head is applied to solve resource allocation with continuous variables.

\item  \emph{Progressive scheduler}: The progressive scheduler is presented to adjust the multi-task agent and the replay buffer pool, which has the capability to leverage the stored policy knowledge by selectively adding or removing specific neurons of the agent, thereby accommodating variations in the number of UAVs without necessitating retraining of the agent. This progressive structure can naturally accumulate experiences for varying number of UAVs and immune to catastrophic forgetting.
 
\item  \emph{Action refinement}: The action refinement is introduced into the proposed DRL for performance enhancement and learning acceleration, and a light taboo search (LTS) guided by channel gains is presented as the action refinement module to balance the exploration and exploitation of the DRL and accelerate optimal policy search in dynamic action space.
\end{enumerate}	

The remainder of this paper can be organized as follows. A review of related work is presented in Section II. The system model and problem formulation are described in Section III. The detailed designs of the FRES framework are introduced in Section IV. The numerical results are presented in Section V, followed by the conclusions in Section VI.

\section{Related Work}
\subsection{UAV-assisted MEC Systems}

In \cite{RN5}, the authors proposed the MEC system with multiple ground servers and an air server. In order to achieve the maximum calculation rate, the author optimized the trajectory of the UAV and the offloading of each user task. The work in \cite{RN6} introduced a single UAV-assisted MEC system, in which the UAV was used to collect the data of users on the ground, and the energy consumed by the UAV was minimized by optimizing the trajectory of the UAV. The work in \cite{RN7} considered a framework in which UAVs were used to assist users in task processing. In order to make all users consume the lowest energy when processing tasks, the authors optimized the trajectory and allocated resources of the UAV. The work in \cite{RN8} proposed a UAV-assisted MEC system based on DRL. The users offloaded the tasks that need to be processed in the UAV to improve the computing efficiency. This method realized the offloading decision and resource allocation under the time-varying channel. The work in \cite{RN9} proposed a MEC system containing multiple UAVs and the number of UAVs was reduced as much as possible while meeting the constraints. In this work, the system energy consumption was minimized by optimizing the offloading decision and resource allocation of user tasks.

\subsection{IRS-assisted UAV Communications}

In\cite{RN13}, in order to improve the quality of the transmission channel, a low-altitude passive relay system was established using IRS-assisted UAV to convert the Rayleigh fading channel into the Rician fading channel to improve the transmission performance. In \cite{RN14}, the author proposed a radio system in which the UAV's signal was reflected to the base station (BS) by IRSs, and the weighted sum rate of all IRSs was maximized by optimizing the trajectory of the UAV and the phase shift matrix of the IRSs. In \cite{RN15}, the author proposed a multi-IRS-assisted UAV system, which improved the user's received power by continuously adjusting the trajectory of the UAV and the beamforming of the IRS. In \cite{RN16}, with the help of the IRS component, the author improved the communication quality between the UAV and the users by adjusting the trajectory of the drone and the phase shift matrix of the IRS. An IRS-assisted UAV communication system was studied in \cite{RN17}. In order to increase the average transmission rate of the system, the phase shift matrix of the IRS and the trajectory of the UAV were jointly optimized.

\subsection{Resource Scheduling in MEC Systems}

At present, there are many related works that formulated the resource scheduling in the MEC system as a MINLP problem, and adopted some traditional solutions to solve it. In \cite{RN18}, the author proposed a dynamic programming algorithm to allocate bandwidth and computing resources for mobile devices to minimize the energy consumption. 
In \cite{RN6}, the successive convex approximation (SCA) algorithm was used to optimize the resource allocation, and the energy efficiency of the UAV in the MEC system is maximized. In \cite{RN19}, the block coordinate descent method and the successive convex approximation algorithm were used to optimize the resource scheduling problem in the cellular connected UAV system. In \cite{RN20}, the author proposed a heuristic algorithm named GAWOA on the basis of genetic algorithm (GA) and whale optimization algorithm (WOA) to solve the resource scheduling problem in the MEC system. In \cite{RN21}, a hierarchical GA and particle swarm optimization (PSO) based computing algorithm was designed to solve the resource scheduling problem in the MEC system.

There are also some related works that used deep learning to implement resource scheduling in the MEC system. In \cite{RN22}, a DRL-based resource scheduling method was proposed. The author reduced the state space of DRL by compressing channel quality information, and proposed adaptive iteration to improve the search efficiency in the DRL. In \cite{RN23}, a resource scheduling framework based on distributed network structure was proposed. By optimized the resource scheduling in the large-scale MEC system, the sum of user energy consumption and delay in the large-scale MEC system was minimized. In \cite{RN24}, a hybrid mobile edge computing (H-MEC) resource scheduling algorithm based on deep learning was proposed, in which a deep neural network with scheduling layers was introduced to realize resource scheduling in the MEC system. 

In Table \ref{tab1}, we compare our FRES framework with existing works, and the comparison consists of IRS, UAV, MEC and deep learning (DL) and reinforcement learning (RL). It can be witnessed that most works listed above consider the system model from two or three perspectives. 
Nevertheless, the aforementioned studies did not specifically investigate the potential of utilizing IRS and UAV-assisted MEC systems in scenarios where the number of UAVs is variable. DRL is a workable method in dynamic environments. Consequently, we introduce DRL with progressive learning to enable real-time resource allocation in dynamic environments.

\begin{table}
	\centering\makegapedcells
	\caption{Comparison with previous works.}
	\label{tab1}
	\begin{tabular}{|p{37pt}<{\centering}|p{37pt}<{\centering}|p{37pt}<{\centering}|p{37pt}<{\centering}|p{37pt}<{\centering}|}

	\hline
Work	& IRS & UAV & MEC & DL\&RL \\
	\hline
[6]-[10]	&  &  \checkmark  &  \checkmark & \\
	\hline
[11]-[15]	& \checkmark & \checkmark & & \\
	\hline
[16]-[19]	&  &  & \checkmark &\\
	\hline
[20]-[22]	&  &  & \checkmark &\checkmark\\
	\hline
FRES	& \checkmark & \checkmark &\checkmark &\checkmark \\
	\hline
\end{tabular}
\end{table}

\section{System Model and Problem Formulation}
\subsection{IRS and UAV-assisted MEC system}
\begin{figure}[htpb]
    \centering
	\includegraphics[width=8.8cm]{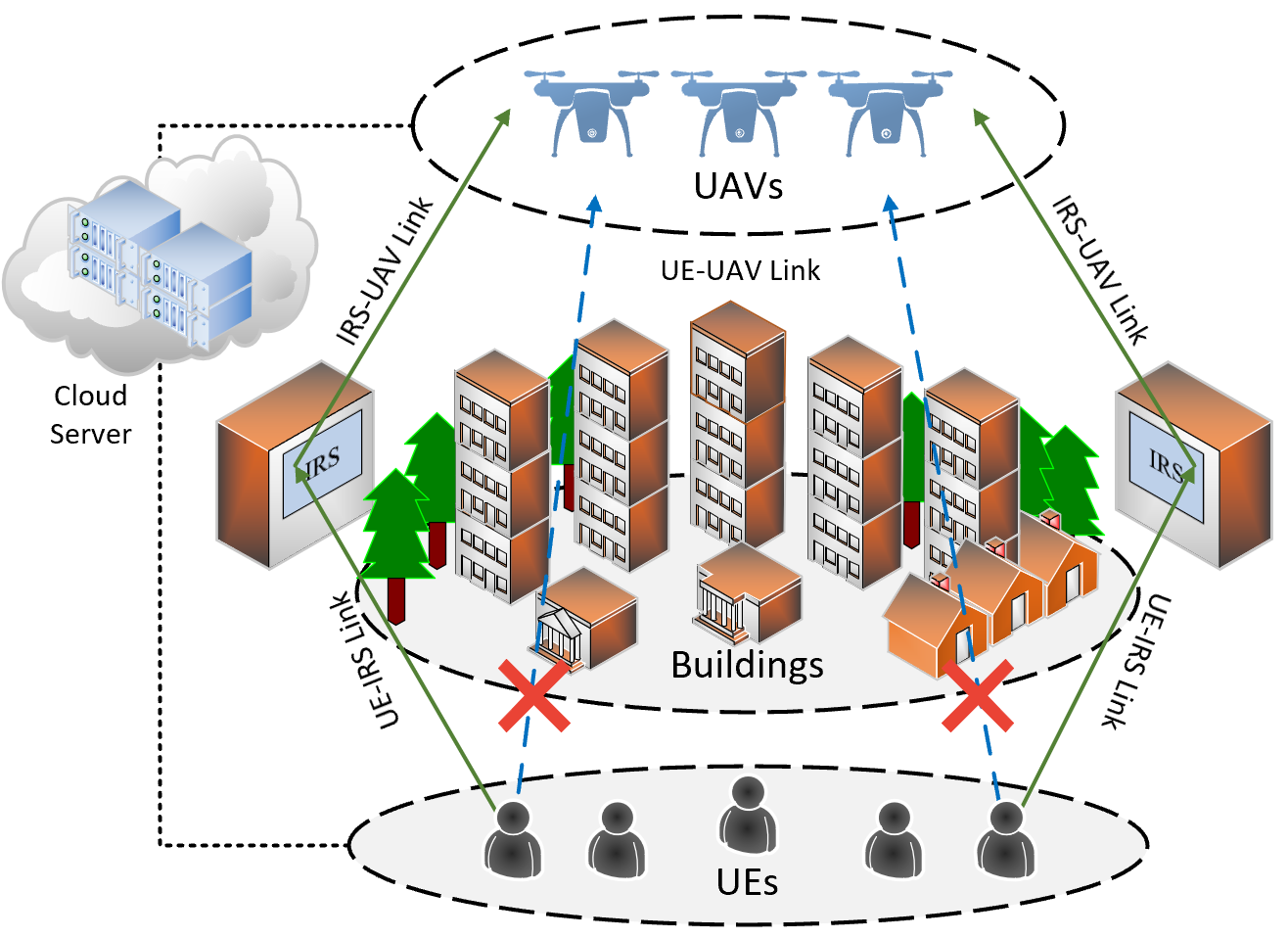}
	\caption{IRS and UAV-assisted MEC system.}
    \label{fig:1}
\end{figure}

We consider the IRS and UAV-assisted MEC system, as shown in Fig. \ref{fig:1}, which consists of $N$ UEs,  $L$ IRSs, some UAVs and a central cloud. All UEs are randomly distributed and all IRSs are mounted on buildings. The set of UEs is denoted as $\mathcal{N}=\{1,2,\ldots,N\}$, the set of UAVs is denoted as $\mathcal{M}=\{1,2,\ldots,M\}$ and the set of IRSs is denoted as $\mathcal{L}=\{1,2,\ldots,L\}$. The central cloud is utilized to gather all system information, encompassing channel state information, UAV information, IRS information, and task-related information from UEs. By deploying the DRL model on the central cloud, the task offloading and resource scheduling for the MEC system are executed.

Each UE has the option to either offload its entire task to a single UAV or execute it locally. We employ the notation $a_{i0}$ to indicate that a task is executed locally, and $a_{ij}$ to signify that a task is offloaded to the $j$th UAV. Consequently, we can establish the following constraints:
\begin{align}
	a_{i0} = \left\{ {0,1} \right\},~~~\forall i \in \mathcal{N}
	\label{C1}
\end{align}
\begin{align}
	a_{ij} = \left\{ 0,1 \right\},~~~\forall i \in \mathcal{N},~\forall j \in \mathcal{M}
	\label{C2}
\end{align}
where $a_{i0}= 1$ signifies that the $i$th UE chooses to execute the task locally. Otherwise, $a_{i0}= 0$. Similarly, $a_{ij} = 1$ implies that the $i$th UE wants to offload its task to the $j$th UAV, while $a_{ij}= 0$ indicates otherwise. Each UE can choose only one location for task execution, we are led to establish the following constraint:
\begin{align}
	a_{i0} + {\sum\limits_{j \in \mathcal{M}}a_{ij}} = 1,~~~\forall i \in \mathcal{N}.
	\label{C3}
\end{align}

We assume that each UE has a computationally intensive task $U_{i}$ which can be expressed as follows:
\begin{align}
U_{i} = \left( {R_{i},F_{i}} \right),~~~\forall i \in \mathcal{N}
\end{align}
where $R_{i}$ denotes the volume of data that is required to be transmitted to the UAV for task execution, while $F_{i}$ signifies the cumulative number of required CPU cycles for task computation.

\subsection{Communication Model}
We assume that UEs are situated in a bustling urban center with numerous tall buildings. In such an environment, the direct link to the UAVs may be obstructed, leading to severe path loss. However, by deploying IRSs on these building surfaces, we can enhance the communication quality between UEs and UAVs.
Each IRS is assumed to have multiple reflecting elements, denoted as $ \mathcal{K} = \left\{ 1,2,\ldots,K \right\} $, and only one IRS is utilized to assist each UE in communication.
The coordinates of the $i$th UE can be represented as  $ \left( x_{i}^{U},y_{i}^{U},z_{i}^{U} \right) $; those of the $l$th IRS can be represented as $ \left (x_{l}^{R},y_{l}^{R},z_{l}^{R} \right) $; and those of the $j$th UAV can be represented as $ \left(x_{j}^{V},y_{j}^{V},z_{j}^{V}\right) $. 
Therefore, we calculate the distance between the  $ i $th UE and the $ l $th IRS as follows:
\begin{align}
	d_{i,l}^{UR} = \sqrt{\left( x_{i}^{U} - x_{l}^{R} \right)^{2} + \left( y_{i}^{U} - y_{l}^{R} \right)^{2} + \left( z_{i}^{U} - z_{l}^{R} \right)^{2}}.
\label{eq1}
\end{align}

Similarly, the distance between the $l$th IRS and $j$th UAV  is expressed as
\begin{align}
d_{l,j}^{RV} = \sqrt{\left( x_{j}^{R} - x_{l}^{V} \right)^{2} + \left( y_{j}^{R} - y_{l}^{V} \right)^{2} + \left( z_{j}^{R} - z_{l}^{V} \right)^{2}}.
\label{eq2}
\end{align}

We consider that the communication path between the $i$th UE and the $j$th UAV is divided into two segments: the UE-IRS link and the IRS-UAV link.
We also assume that each UE only uses one IRS for signal transmission, so that the overall channel gain $ h_{i}^{UR} $ of the UE-IRS link for the $ i $th UE can be expressed as $ {h}_{i,l}^{UR} $. Therefore, we can calculate   $ {h}_{i,l}^{UR} $ as follows:
\begin{align}
	h_{i,l}^{UR} = \sqrt{\frac{\epsilon}{\left( d_{i,l}^{UR} \right)^{\alpha}}}\left\lbrack {1,e^{- j\frac{2\pi}{\lambda}d\phi_{i,l}^{UR}},\ldots,e^{- j\frac{2\pi}{\lambda}{\lbrack{K - 1}\rbrack}d\phi_{i,l}^{UR}}} \right\rbrack^{T}
	\label{eq:4}
\end{align}
where $ \epsilon $  is the path loss at the reference distance of one meter, $ \alpha $ is the path loss exponent of the channel between the $ i $th UE and the $ l $th IRS, $ \phi_{i,l}^{UR} = \frac{\left| X_{i}^{U} - X_{l}^{R} \right|}{d_{i,l}^{U,R}} $ is the cosine value of the angle of departure (AOD) of the signal between the $ i $th UE and the $ l $th IRS.

Then, we assume that each UE can offload its task to a single UAV only. Consequently, for the $i$th UE, the overall channel gain of the IRS-UAV link, denoted as $ h_{i}^{RV} $, can be represented by $ {h}_{l,j}^{RV} $. Therefore, we can express  $ h_{l,j}^{RV} $ as follows:
\begin{align}
h_{l,j}^{RV} = \sqrt{\frac{\epsilon}{\left( d_{l,j}^{RV} \right)^{2}}}\left\lbrack {1,e^{- j\frac{2\pi}{\lambda}d\phi_{l,j}^{RV}},\ldots,e^{- j\frac{2\pi}{\lambda}{\lbrack{K - 1}\rbrack}d\phi_{l,j}^{RV}}} \right\rbrack^{T}
\label{eq:3}
\end{align}

The right term in Eq. (\ref{eq:3}) above is the array response of the $ l $th IRS which have $ K $ reflecting elements, $ d $ is the antenna separation distance between two elements, $ \phi_{l,j}^{RV} = \frac{\left. \left| X \right._{l}^{R} - X_{j}^{V} \right|}{d_{l,j}^{RV}} $ is the cosine value of the angle of arrival (AOA) of the signal between the $ j $th UAV and the $ l $th IRS, $\lambda$ is the carrier wavelength.

We consider that each IRS contains $ K $ elements, and the phase shift of the $k$th reflecting element of the $l$th IRS is represented by $ \theta_{k,i,l,j} \in \left\lbrack 0,2\pi \right) $.
As a result, we can define a phase-shifted diagonal matrix for IRS-assisted signal transmission as follows:
\begin{align}
\Theta_{i,l,j} = \text{diag}\left\{ e^{j\theta_{k,i,l,j}},\forall k \in \mathcal{K} \right\}.
\label{QPB2}
\end{align}

In our assumptions, UEs offload their tasks to UAVs through orthogonal frequency division multiplexing (OFDM) channels, thereby ensuring the absence of interference among the UEs. Hence, when the $i$th UE chooses to offload its task to the $j$th UAV via the $l$th IRS, the achieved data transmission rate can be mathematically formulated as follows:
\begin{align}
	r_{ij} = {{B\log}_{2}\left( {1 + \frac{P_{ij}^{t}\left| (h_{i,l}^{UR} )^{T}\Theta_{i,l,j}(h_{l,j}^{RV})
			 \right|^{2}}{\sigma^{2}}} \right)}
	\label{eq9}
\end{align}
where $ B $ denotes the channel bandwidth, $ \sigma^{2} $ signifies the noise spectral density, $ P_{ij}^{t} $ represents the transmitting power between the $ i $th UE and the $ j $th UAV. 

The energy consumption of transmission can be calculated as follows:
\begin{align}
	E_{ij}^{t} = P_{ij}^{t}\frac{R_{i}}{r_{ij}}.
	\label{eq11}
\end{align}

\subsection{Computation Model}
\subsubsection{Local Computation Model}

The energy consumption of the local computing can be expressed as follows:
\begin{align}
	E_{i}^{l} =\nu_{1}\left( f_{i0} \right)^{\tau_{1}-1} F_{i},~~~\forall i \in \mathcal{N} 
	\label{eq14}
\end{align}
where $f_{i0}$ is the local computation capacity of the $i$th UE. $ \nu_{1} \geq 0 $ is the effective switched capacitance and $ \tau_{1} \geq 1 $ is the positive constant. 

The computation capacity of the $ i $th UE is constrained by:
\begin{align}
	a_{i0}f_{i0} \leq F_{i,max}^{L},~~~~\forall i \in \mathcal{N}
	\label{C4}
\end{align}
where $ F_{i,max}^{L} $ means the maximum local computation capacity of the $ i $th UE.

\subsubsection{Remote Computation Model}
The energy consumption in remote computing phase is
\begin{align}
	E_{ij}^{u} = \nu_{2}\left( f_{ij} \right)^{\tau_{2}-1}F_i
	\label{eq17}
\end{align}
where $ f_{ij} $ means that the computing resource allocated by the $ j $th UAV to the $ i $th UE. $ \nu_{2} $  is the effective switched capacitance and $ \tau_2 \geq 1 $ is the positive constant. 

Since each UAV has a limited computation capacity, we can represent the constrained computation resource of the $j$th UAV as follows:
\begin{align}
	{\sum\limits_{i \in \mathcal{N}}{a_{ij}f_{ij}}} \leq F_{j,max}^{R},~~~~\forall j \in \mathcal{M}
	\label{C5}
\end{align}
where $ F_{j,max}^{R} $ is the total computation capacity of the $ j $th UAV.

\subsection{UAV Hover Model}

When the $ j $th UAV hovers to receive and perform offloaded tasks, the transmission time of the offloaded task can be expressed as:
\begin{align}
	T_{ij}^{t} = \frac{R_{i}}{r_{ij}},~~~\forall i \in \mathcal{N},\forall j \in \mathcal{M}.
	\label{eq10}
\end{align}

The execution time of the offloaded task can be expressed as:
\begin{align}
	T_{ij}^{e} = \frac{F_{i}}{f_{ij}},~~~\forall i \in \mathcal{N},\forall j \in \mathcal{M}.
	\label{eq10}
\end{align}

Hence, the energy consumption associated with UAV hovering is given by
\begin{align}
	E_{j}^{h}=P_{j}^{h}\cdot \max \limits_{i \in \mathcal{N}} \{a_{ij}\left(T_{ij}^{t}+T_{ij}^{e}\right)\},~~~\forall j \in \mathcal{M}
\end{align}
where $ P_{j}^{h} $ means the hover power of the $ j $th UAV.

\subsection{Problem Formulation}
In this study, our objective is to minimize the energy consumption of the IRS and UAV-assisted MEC system. The optimization problem can be summarized as follows:

\begin{equation*}
\begin{aligned}
	P0:\min\limits_{\mathbf{L},\mathbf{A},\mathbf{F},\mathbf{\Theta}}{\sum\limits_{i \in \mathcal{N}}{\left( {a_{i0}E_i^{l}+\sum\limits_{j \in \mathcal{M}}{a_{ij}E_{ij}^{t}}}\right)}}
	\\
	+ {\sum\limits_{j \in \mathcal{M}}{\left(\lambda E_{j}^{h}+\sum\limits_{i \in \mathcal{N}}{a_{ij}E_{ij}^{u}} \right)}}
\end{aligned}
\end{equation*}
\begin{equation}
	\text {s.t.} ~(\ref{C1}), (\ref{C2}),(\ref{C3}),(\ref{C4}),(\ref{C5}).
	\label{eq:P0}
\end{equation}
where $\mathbf{L}=\left\{ x_{j}^{V},y_{j}^{V},z_{j}^{V} \middle| j \in \mathcal{M} \right\}$ denotes the positions of UAVs. The offloading decision is represented by $ \mathbf{A} = \left\{a_{i0},a_{ij} \middle| i \in \mathcal{N},j \in \mathcal{M}\right\}$, while the allocation of computing resources is expressed as $\mathbf{F}=  \left\{f_{i0},f_{ij}\middle| i\in\mathcal {N}, j\in\mathcal {M}\right\}$.
The phase-shifted diagonal matrix for IRS-assisted signal transmission is denoted by  $ \mathbf{\Theta} = \left\{ \Theta_{i,l,j} \middle| i \in \mathcal{N}, l \in \mathcal{L}, j \in \mathcal{M} \right\} $.
Lastly, $\lambda$ serves as a weight coefficient. For saving flight energy, we assume that UAVs need to be redeployed only when the number of UAVs is changed. We use large-scale path-loss fuzzy c-means clustering algorithm (LS-FCM) to optimize the locations of UAVs and obtain the $ \mathbf{L} $ in \textit{P}0 \cite{RN24}.

\section{The FRES Framework}

Problem \textit{P}0 is a typical MINLP, where $ \mathbf{A} $ are binary variables, and $ \mathbf{L},~\mathbf{F} $ and $ \mathbf{\Theta} $ are real positive variables. This makes the Problem \textit{P}0 non-smooth and non-differential. Moreover, there are still three challenges for solving Problem \textit{P}0: (1) It is a large-scale optimization problem with many variables, the search is easy to trap into the local minimum. (2) In the system, the number of UAVs is varying, so the prior knowledge is hard to be reused to solve the problem in the dynamic environment. (3) It needs to design a real-time resource scheduling with low computational complexity for the wireless fading channel.

To this end, we propose a deep progressive reinforcement learning based FRES framework to address the above challenges. First, we present a model free DRL-based resource scheduling structure, which can learn from the environment by the interaction between the agent and the MEC system without any prior information. After the agent of DRL is well trained, the agent can make real-time decision easily by doing some simple algebraic calculations instead of solving the original \textit{P}0 repeatedly. Second, we propose a multi-task agent with two-head structure to solve the MINLP directly. Third, we consider a progressive scheduler to adjust structure of the agent for adapting the varying number of UAVs, which can add or remove a part of neurons in the agent to trace the dynamic environment. This structure is resistant to  catastrophic forgetting. Finally, we introduce an action refinement to enhance the exploration, in which a LTS guided by channel gains is designed to jump out of the local minimum and accelerate the policy search in large action space.

The dataflow of the FRES framework can be described as follows: first, the central cloud collects the global environment information as well as task information from the MEC system. Then, the central cloud executes the FRES algorithm, updates the locations of UAVs and the phase-shifted diagonal matrix of IRSs, and performs the online decisions of the user association and resource allocation for each UE. Finally, based on the decision received from the central cloud, each UE can offload the task to the suitable UAV, and then receive results accordingly. 

\begin{figure*}[h]
	\centering
	\includegraphics[width=13cm]{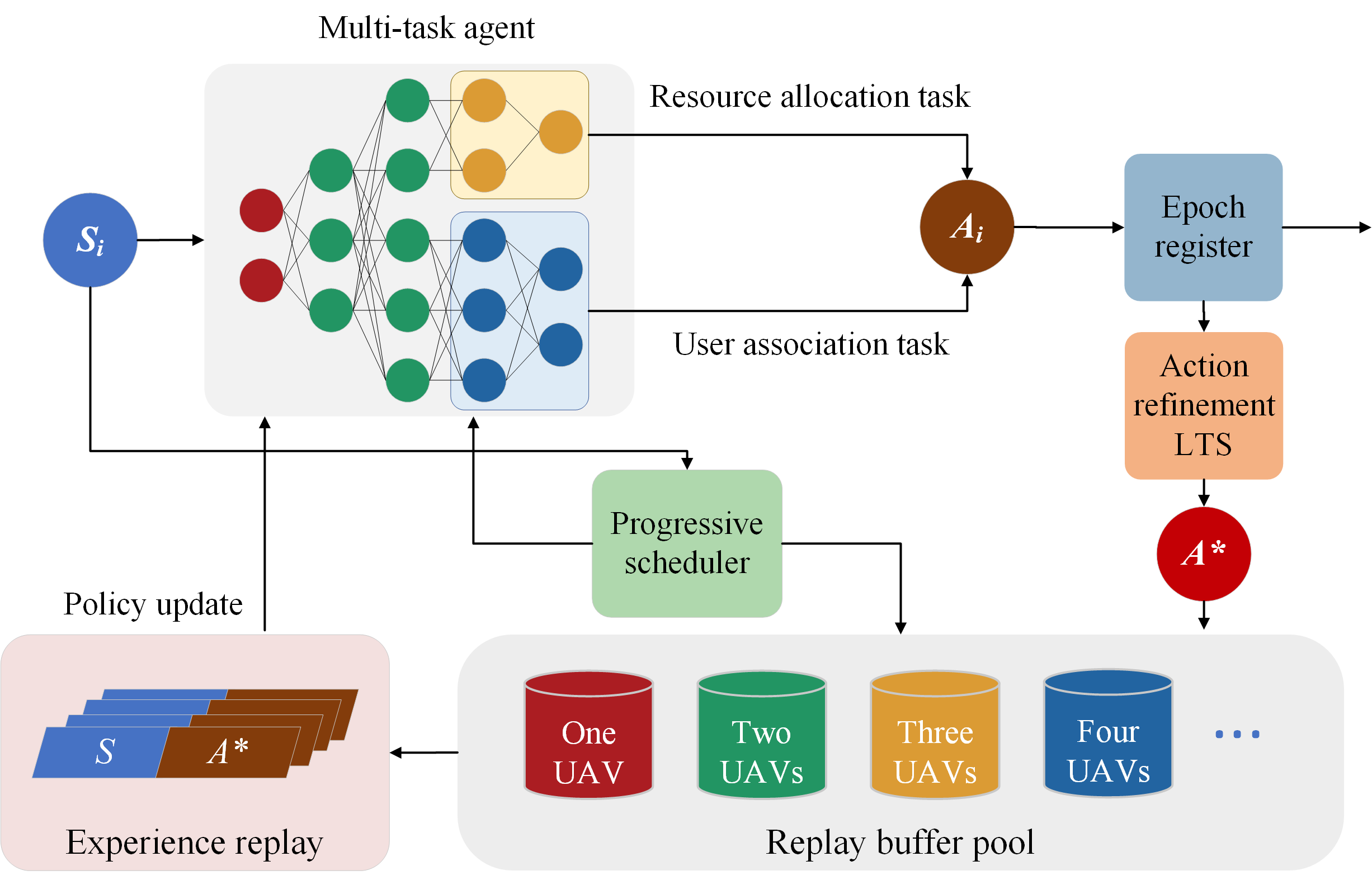}
	\caption{The resource scheduling part of the FRES framework.}
	\label{fig:2}
\end{figure*}

\subsection{Algorithm Overview}
The details of the FRES framework are described in \textbf{Algorithm 1}. We first introduce a quantitative passive beamforming (QPB) method to solve  $ \mathbf{\Theta}_t $ according to the positions of UAVs and UEs. Then, we design a model-free DRL to generate  $ \mathbf{A}_t $ and $ \mathbf{F}_t $ of all UEs at $ t $th timeslot.
Next, we introduce the state, action, and reward of the DRL in the FRES framework as follows:

\begin{algorithm}[h]
	\caption{FRES framework}
	\label{alg1}
	\begin{algorithmic}[1]
		\REQUIRE   $R_{i,t},F_{i,t}$.
		\ENSURE $\mathbf{L}_t,\mathbf{A}_t,\mathbf{F}_t,\mathbf{\Theta}_t$. 
		\STATE{Initialize a replay buffer pool $\mathcal{P}$.}\\
		\STATE{Initialize the multi-task agent's parameters $ \mathbf{w}_{m} $ for $m$ UAVs randomly.}
		
		$\textbf{Online\;inference\;stage}$\\
		\WHILE{$t<T_{OL}$} 
		\IF{there is a variation in the number of UAVs}
		\STATE{Modify the parameters $ \mathbf{w}_{m,t} $ of the multi-task agent and choose the appropriate replay buffer $R_{m,t}$ from the replay buffer pool $\mathcal{P}$ by \textbf{Algorithm \ref{alg3}}.}
		\STATE{Calculate the locations $\mathbf{L}_t$ of UAVs.}
		\STATE{Calculate the phase-shifted diagonal matrix $\mathbf{\Theta}_t$ by  Eqs. (\ref{QPB1}) and (\ref{QPB2}).}
		\ENDIF
		\FOR{$ i = 1,\ldots,N $}
		\STATE{Generate action $\mathcal{A}_{i,t} = \pi\left( \mathcal{S}_{i,t} \middle|  \mathbf{w}_{m,t}  \right)$ by the forward process of \textbf{Algorithm\enspace\ref{alg2}}.}
		\ENDFOR
		\STATE{Check the constraints for the action of all UEs $ \mathcal{A}_{t} = \left\{ \mathcal{A}_{i,t}\middle| i \in \mathcal{N} \right\} $ and evaluate the current solution $ \mathcal{A}_{t} $.} 
		$\textbf{Offline\;learning\;stage}$\\		
		\IF{constraints cannot be guaranteed}
		\STATE{Seek a near-optimal action $ \mathcal{A}_{t}^{*} $ from the initial action $ \mathcal{A}_{t} $ as described in\textbf{ Algorithm \ref{alg4}}.}
		\STATE{Append the state-action transition pair $\left\{{\mathcal{S}_{i,t},\mathcal{A}_{i,t}^{*}}\right\}$ to the current replay buffer $R_{m,t}$.}
		\STATE{Sample a batch of transitions and feed them to the multi-task agent.}
		\STATE{Update the multi-task agent's parameters $\mathbf{w}_{m,t}$ according to the backward process described in \textbf{Algorithm \ref{alg2}}.}
		\ENDIF
		\ENDWHILE
	\end{algorithmic}
\end{algorithm}

\subsubsection{State} $ \mathcal{S}_{i,t} = \left\{ H_{i,t},R_{i,t},F_{i,t} \right\} $ is defined as the state information of the $ i $th UE at $ t $th timeslot, in which $ H_{i,t} = \left\{ h_{i,j,t} \middle| j \in \mathcal{M} \right\} $ is the channel gain between the $ i $th UE and all UAVs at $ t $th timeslot. 

\subsubsection{Action} 
$ \mathcal{A}_{i,t} = \left\{ a_{i,t},f_{i,t} \right\} $ is defined as the resource scheduling decision of the $ i $th UE at $ t $th timeslot, in which $ a_{i,t}$ and $ f_{i,t}$ are the user association and the allocated resource of the $ i $th UE, respectively. 

\subsubsection{Reward} 
$ \mathcal{R}_{t} $ is defined as the reciprocal of the total energy consumption of the IRS and UAV-assisted MEC system (The goal of Problem \textit{P}0).


Then, the structure of the resource scheduling part of the FRES framework is illustrated in  Fig. \ref{fig:2}. Firstly, we build a multi-UAVs and multi-UEs environment, and we initialize the parameters $ \mathbf{w}_{m} $ of the multi-task agent randomly. We also initialize an empty replay buffer pool $ \mathcal{P} $ which includes many replay buffers for progressive scheduler. Then, we collect the environment information $ \mathcal{S}_{i,t} $ of the $ i $th UE at $ t $ timeslot, Next, $ \mathcal{S}_{i,t} $ is fed to the multi-task agent and solve the MINLP by the forward process of \textbf{Algorithm 2}, and the action $ \mathcal{A}_{i,t} $ is generated by the multi-task agent. All the UE's $ \mathcal{A}_{t} = \left\{ \mathcal{A}_{i,t} \middle| i \in \mathcal{N} \right\} $ at $ t $ timeslot are stored in an epoch register \cite{jiang2020ai}, in which we can check whether the constraints are met at this epoch. If the constraints cannot be guaranteed, the current $ \mathcal{A}_{t} $ will be sent to the action exploration module to search for a better solution. \textbf{Algorithm 4} is applied as the action exploration method to find a near-optimal action $ \mathcal{A}_{t}^{*} $ for all UEs. The new transition $ \left\{ \mathcal{S}_{i,t},\mathcal{A}_{i,t}^{*} \right\} $ is added to the replay buffer $R_{m,t}$ until it is full, and then the first in first out (FIFO) strategy is introduced to over-write old data. Prioritization experience replay is used to sample a batch of transitions, and the parameters $ \mathbf{w}_{m,t} $ of the multi-task agent are updated by the backward process of the \textbf{Algorithm 2}. Moreover, when the number of the UAVs changes, the progressive scheduler will start work, and the structure of the multi-task agent will be adjusted to adapt to the new environment and the corresponding replay buffer will be selected by \textbf{Algorithm 3}.

\subsection{QPB}
We introduce the QPB method to optimize the phase shift matrix of IRSs \cite{dong2022joint}. Due to hardware facility restrictions, the IRS can only reflect signals with certain phase shifts. For simplicity, we consider discrete phase shift angles in this paper, the phase shift $ \theta_{k,i,l,j} $ of the IRS is chosen from the following set of $ \Psi \triangleq \left\{ {\frac{2\pi}{N_{p}}i,i = 0,1,\ldots,N_{p} - 1} \right\}$, where $ N_{p} $ denotes the number of the phase shift values that can be selected for every element. We optimize the phase shift $ \theta_{k,i,l,j} $ of the $ k $th reflecting element in the $ l $th IRS between the $ i $th UE and the $ j $th UAV with the following equation:
\begin{align}
	\theta_{k,i,l,j} = {\underset{\theta_{k,i,l,j}^{'} \in \Psi}{\text{argmin}}\left| {\theta_{k,i,l,j}^{'} - \left( {\omega_{k,i,l}^{UR}} + \omega_{k,l,j}^{RV} \right)} \right|}
	\label{QPB1}
\end{align}
where  $ \omega_{k,i,l}^{UR} \in \left\lbrack 0,2\pi \right) $ denotes the phase shift of the $ k $th reflecting element from the $ i $th UE to the $ l $th IRS, and $ \omega_{k,l,j}^{RV} \in \left\lbrack 0,2\pi \right) $ denotes the phase shift of the $ k $th reflecting element from the $ l $th IRS to the $ j $th UAV.

\subsection{Multi-task Agent}
\begin{figure}[h]
	\centering
	\includegraphics[width=9.5cm]{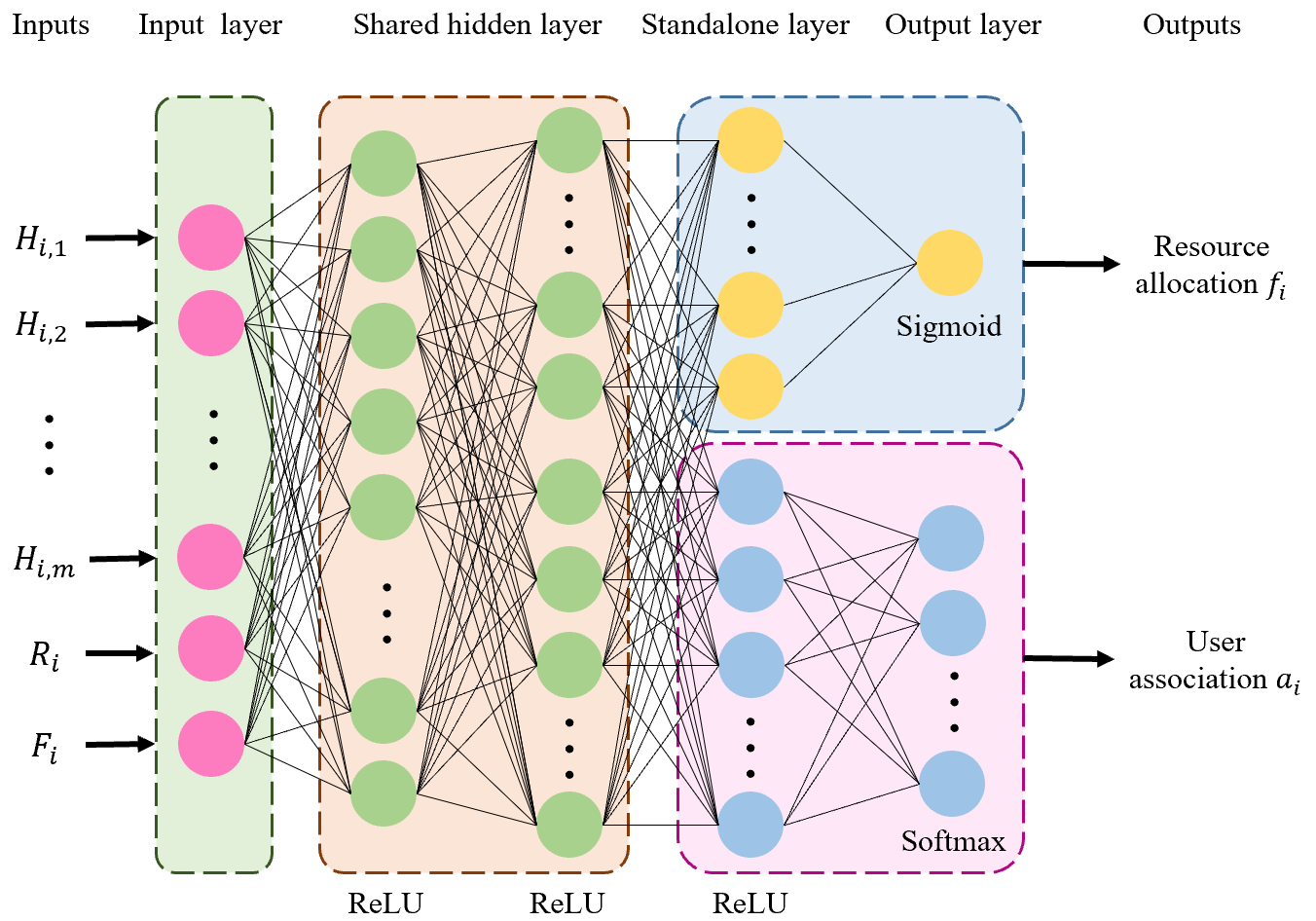}
	\caption{The structure of multi-task agent.}
	\label{fig:3}
\end{figure}

Multi-task learning is a machine learning method that puts multiple related tasks together to learn\cite{10040995}. In the learning process, these tasks share parameters of the shallow part of the deep neural network (DNN) to extract the common features and then adjust the independent parameters of the subsequent part of the DNN to learning the unique features of each task respectively. The whole multi-task agent includes four part: input layer, shared layer, standalone layer and output layer, which are shown in Fig. \ref{fig:3}.

As depicted in Fig. \ref{fig:3}, the multi-task agent consists of $Q$ shared layers, and $P$ standalone layers for each specific task.
The output from the $q$th shared layer can be characterized as follows:
\begin{align}
	\mathbf{o}_{q} = \sigma\left( \mathbf{w}_{q}\mathbf{o}_{q - 1} + \mathbf{b}_{q} \right)
	\label{eq25}
\end{align}
where $ \mathbf{w}_{q} $ and  $ \mathbf{b}_{q}$ represent the weights and biases of the $q$th shared layer, respectively. $\sigma(.)$ is the activation function.

We can express the output from the $p$th standalone layer of the $j$th task as follows:
\begin{align}
	\mathbf{o}_{j,Q + p} = \sigma \left( \mathbf{w}_{j,Q + p}\mathbf{o}_{j,Q + p - 1} + \mathbf{b}_{j,Q + p} \right)
	\label{eq26}.
\end{align}
where $ \mathbf{w}_{j,Q + p} $ and $\mathbf{b}_{j,Q + p}$ represent the weights and biases of the $p$th standalone layer of the $j$th task, respectively. 

The output of the $P$th standalone layer of the $j$th task can be articulated as follows:
\begin{align}
	\mathbf{o}_{j,Q + P} = \left\{ \begin{matrix}
		{\text{Sigmoid}\left( \mathbf{w}_{\textit{j,Q + P}}\mathbf{o}_{j,Q + P - 1} + \mathbf{b}_{j,Q + P} \right),}\\ \text{if~it~is~the~resource~allocation.} \\
		{\text{Softmax}\left( \mathbf{w}_{j,Q + P}\mathbf{o}_{j,Q + P - 1} + \mathbf{b}_{j,Q + P} \right),}\\
		\text{if~it~is~the~user~association.} \\
	\end{matrix} \right.
	\label{eq27}
\end{align}
where $ \mathbf{w}_{j,L + K} $ and $ \mathbf{b}_{j,L+K} $ represent the weights and biases of the output layer of the $ j $th task, respectively. Sigmoid and Softmax are the activation functions of the output layer.

In our study, the offloading decision of each UE is generally an integer, and the resource allocation of each UE is usually a real number. Hence, we decompose the original MINLP problem into two ML tasks: the classification task with user association and the regression task with resource allocation. Then, we design a multi-task DNN, in which the shared layers are used to learn the common features of the MINLP problem, a regression head with standalone layers is applied to solve the resource allocation task, and a classification head with standalone layers is applied to solve the user association task. Additionally, ReLU function is used in the hidden layers, sigmoid function is used in the output of the regression head, and softmax function is used in the output of the classification head. 

In the user association head, the loss function employs cross-entropy loss, which can be formulated as follows:
\begin{align}
	L_{ce} = \frac{1}{S}{\sum\limits_{k}^{S}{\left( {- {\sum\limits_{j = 1}^{M}{y_{kj}{\log p_{kj}}}}} \right)}} 
	\label{eq28}
\end{align}
where $S$ is the number of selected transitions, $ y_{kj} $ is an indicator variable, $ y_{kj} = 1 $ means that the true label is same as the predicted label, and $ p_{kj} $ denotes the probability of the $ k $th UE offloading to the $ j $th UAV. 

In the resource allocation head, the loss function utilizes mean square error (MSE) loss, which can be calculated as follows:
\begin{align}
	L_{mse} = \frac{1}{S}{\sum\limits_{k}^{S}\left( {\hat{y_{k}} - y_{k}} \right)^{2}} 
	\label{eq29}
\end{align}
where $ \hat{y_{k}} $ means the predicted value and $ y_{k} $ means the true value.
The total multi-task loss $ L_{mt} $ can be expressed as
\begin{align}
	L_{mt} =L_{ce} + \xi L_{mse}
	\label{eq30}
\end{align}
where $ \xi $ is the weighted coefficient. The details of the multi-agent are described in \textbf{Algorithm 2}.
\begin{algorithm}[H]
	\caption{Multi-task agent}
	\label{alg2}
	\begin{algorithmic}[1]
		\REQUIRE   $\mathcal{S}_{i,t}$.
		\ENSURE  $\mathcal{A}_{i,t}$, ${\mathbf{w}}_{m,t}$.
		\\$ \textbf{Forward~process} $
		\STATE{Feed $ \mathcal{S}_{i,t} $ into the input layer of the multi-task agent.}
		\STATE{Calculate the outputs of the shared layers by Eq. (\ref{eq25}).}
		\FOR{each task $ j $}
		\STATE{Calculate the outputs of the standalone layers of the $ j $th task by Eq. (\ref{eq26})}.\\
		\STATE{Calculate the output of the $ j $th task by Eq. (\ref{eq27})}.\\
		\ENDFOR\\
		\STATE{Generate the scheduling decision $\mathcal{A}_{i,t}$ from all tasks}.\\
		$\textbf{Backward~process}$\\
		\STATE{Calculate the loss function of the user association head by Eq. (\ref{eq28}).}
		\STATE{Calculate the loss function of the resource allocation head by Eq. (\ref{eq29}).}
		\STATE{Calculate the total multi-task loss function by Eq.  (\ref{eq30}).}
		\FOR{each task $ j $}
		\STATE{Calculate the gradient and update the parameters of the standalone layers of the $ j $th task.}
		\ENDFOR\\
		\STATE{Calculate the gradient of the total multi-task loss function and update the parameters of the shared layers.}
		\STATE{Obtain the whole parameters ${\mathbf{w}}_{m,t}$ from all tasks}.\\
	\end{algorithmic}
\end{algorithm}

\subsection{Progressive Scheduler}
Progressive learning was first proposed by Google DeepMind \cite{PNN}, which was a novel knowledge transfer method based on continual learning\cite{9724647}. With the help of the previous knowledge of the old task, the new task can be trained relatively quickly, and the parameters of the old tasks will not be forgotten. When the old task needs to be carried out again, the new knowledge will be forgotten and the old model parameters can be directly recalled to process it without retraining over again.

\begin{figure*}[ht]
	\centering
	\includegraphics[width=18cm]{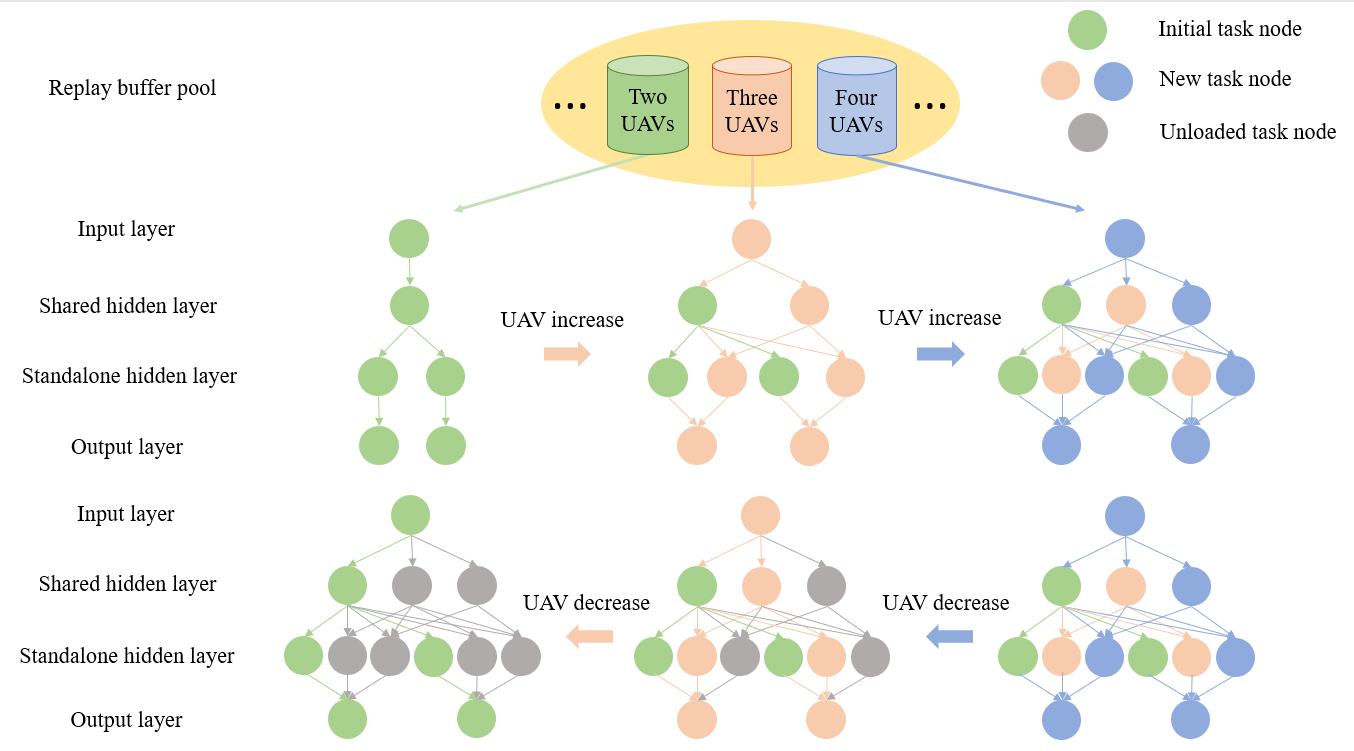}
	\caption{The progressive scheduler incorporating dynamic neuron allocation for UAVs.}
	\label{fig:4}
\end{figure*}

The system model considered in this paper is a dynamic model with different number of UAVs when the environment varies. However, the structure of traditional neural network is stationary, when the number of UAVs changes, the structure of the neural network should be adjusted and the network should be retrained again for adapting the changes in the environment. We design a new progressive scheduler for the multi-task agent to address this challenge, in which we can adjust the structure of the neural network to trace the dynamic environment and we can add or remove a part of the neurons to adapt the change of UAV numbers. The process of dynamically adjusting the neurons using the progressive scheduler for different numbers of UAVs is illustrated in Fig. \ref{fig:4}.

To achieve this goal, we create a replay buffer pool, in which there are several replay buffers in the pool to train the multi-task agent with different structures, then we define the rule of how to adjust the structure of the multi-task agent. The adding rule of UAV increase can be expressed as follows.

The output of the {$q$}th shared layer can be expressed as:
\begin{align}
	\mathbf{o}_{n,q} = \sigma\left( \mathbf{w}_{n,q}\mathbf{o}_{c,q - 1}  + \mathbf{w}_{n,q}\mathbf{o}_{n,q - 1} + \mathbf{b}_{n,q} \right)
	\label{eq31}
\end{align}
where $ c $ means the current neurons and $ n $ means the new neurons.

The output of the $ p $th standalone layer of the $ j $th task can be represented as:
\begin{align}
	\mathbf{o}_{n,j,Q+p} = \sigma\left( \mathbf{w}_{n,j,Q+p}\mathbf{o}_{c,j,Q+p - 1}  \right.
	\notag
	\\ 
	\left.+ \mathbf{w}_{n,j,Q+p}\mathbf{o}_{n,j,Q+p - 1} + \mathbf{b}_{n,j,Q+p} \right).
	\label{eq32}
\end{align}

The removing rule of UAV decrease can be expressed as:
\begin{align}
	\mathbf{w}_{n,q} = 0,\mathbf{b}_{n,q} = 0,\mathbf{w}_{n,j,Q+p} = 0,\mathbf{b}_{n,j,Q+p} = 0.
	\label{eq33}
\end{align}

The details of the progressive scheduler are described in \textbf{Algorithm 3}.

\begin{algorithm}
	\caption{Progressive scheduler}
	\label{alg3}
	\begin{algorithmic}[1]
		\REQUIRE   $\mathcal{S}_{i,t}$.
		\ENSURE ${\mathbf{w}}_{m,t}$, $R_{m,t}$.
		\STATE{Detect the number of the UAVs in the environment according to $\mathcal{S}_{i,t}$.}
		\IF{the number of UAVs increase}
		\STATE{Carry out the adding rule and add the extra neurons to each layers by Eqs. (\ref{eq31})-(\ref{eq32}).}
		\STATE{Select the corresponding replay buffer in the replay buffer pool.}
		\ELSIF{the number of UAVs decrease}
		\STATE{Carry out the removing  rule and turn off the extra neurons of each layers by Eq. (\ref{eq33}).}
		\STATE{Select the corresponding replay buffer in the replay buffer pool.}
		\ELSE
		\STATE{Remain the current structure and replay buffer constantly.}
		\ENDIF
	\end{algorithmic}
\end{algorithm}

\subsection{LTS}
Since the action space of our DRL framework is large and dynamic, the DRL is hard to converge in the action space.
 Action refinement has proven to be an effective global search strategy when dealing with large-scale action spaces in DRL \cite{dulac2015deep}. 
 Based on taboo search\cite{RN26}, we propose an LTS algorithm to implement  action refinement. The details of the LTS are described as follows:
 
\subsubsection{Solution representation}
The solution in taboo search is encoded as a vector $ x = \left( \mathbf{A},\mathbf{F} \right) $, where $ \mathbf{A} $ represents the user association, and $ \mathbf{F} $ denotes the resource allocation. The initial solution $ x_0 = \left( \mathbf{A_0},\mathbf{F_0} \right) $ is acquired from the multi-task agent.

\subsubsection{Evaluation function}
We define the evaluation function of the taboo search as:
\begin{align}
{\text{min}{f\left( x \right)}}: x \in \mathcal{D}
\label{eq34}
\end{align}
where $ f\left( \cdot \right) $ is the minimum energy consumption function in the current scenario which can be expressed as Eq. (\ref{eq:P0}), $ \mathcal{D} $ is the feasible solution space. 
\subsubsection{Neighborhood structure and light move operator}
We generate a neighborhood $ \mathcal{D}_{c} $ of the current solution $ x_{c} $ by changing the offloading decision of some UEs randomly. We then define that the light move operator changes the offloading decision of the selected UE to the place with the highest channel gain. 

\subsubsection{Taboo list}
In taboo search, the recent move operators are stored in the taboo list to avoid some specific move operators which are selected repeatedly and the taboo list is formed with the $ L $ most recently executed move operators. 

\subsubsection{Optimizing strategy} 

When we get the best solution $ x' $ and corresponding move operator $ p' $ in $ \mathcal{D}_{c} $, we check the taboo list firstly. If  $ p' $ is in the taboo list and  $ f\left( x' \right) $ is better than $ f\left( x_{b} \right) $ ($ x_{b} $ is the global best solution), $ x' $ will be selected as $x_c$ and $x_b$ will be updated, else the second best solution $ x'' $ will be selected as the $x_c$ and the corresponding move operator $ p'' $ will be stored into the taboo list. Because  the second best solution $ x'' $ can also be accepted as $x_c$, the optimizing strategy can jump out of the local optimal point.

\subsubsection{Constraint check and guarantee}

We will perform a constraint check on the new solution to ensure that the resources allocated by the UAV to the UEs do not exceed the maximum computational resources of the UAV. If the allocated resources of the UAV are overflowing, we will proportionally reduce the allocated resources for each UE until the constraint is satisfied.

The detailed procedure of LTS is shown in \textbf{Algorithm 4}.

\begin{algorithm}
	\caption{LTS}
	\label{alg4}
	\begin{algorithmic}[1]
		\REQUIRE   $\mathbf{A_0}, \mathbf{F_0}$.
		\ENSURE  $x_b$.
		\STATE{$ x_{0} = \left( \mathbf{A_0},\mathbf{F_0} \right) $, $ x_{c} \leftarrow x_{0} $, $ x_{b} \leftarrow x_{0} $.}
		\STATE{Construct a null taboo list.}
		\WHILE{$ k < T_{max}$}
		\STATE{Generate a neighborhood $ \mathcal{D}_{c} $ around $ x_{c} $ according to the channel state information. }
		\STATE{Find the best solution $ x' $ and the corresponding move operator $ p' $  in $ \mathcal{D}_{c} $ by light move operator.}
		\IF{$ p' $ is in taboo list}
		\IF{$ f\left( x' \right) < f\left( x_{b} \right) $}
		\STATE{$ x_{c}\leftarrow x' $.}
	
		\ELSE
		\STATE{$ x_{c}\leftarrow x'' $.}
		\STATE{Store $ p'' $ into taboo list.}

			\ENDIF
		\ELSE
		\STATE{Store $ p' $ into taboo list.}
				\STATE{$ x_{c}\leftarrow x' $.}
		\ENDIF

		\IF{$ f\left( x_{c} \right) < f\left( x_{b} \right) $}
		\STATE{$ x_{b}\leftarrow x_{c} $.}
		\ENDIF
		
		\STATE{$ k = k+1 $.}
		\ENDWHILE
		\STATE{Carry out the constraint check and obtain a valid solution $x_b$. }
	\end{algorithmic}
\end{algorithm}

\subsection{Time complexity analysis of the FRES framework}

The time complexity of the proposed FRES framework is calculated based on LS-FCM, QPB, the learning of multi-task agent and the LTS. We analyse the time complexity of each part, then compute the overall time complexity of the FRES framework.

The time complexity of LS-FCM is $O(M)$ \cite{RN24}, where $M$ is the current number of UAVs. The time complexity of QPB method is also $O(M)$. The time complexity of multi-task agent learning is $O( M \times T_N)$, where $T_N$ is the iteration number of multi-task agent learning. The time
complexity of LTS is $O(M \times {T_{max}})$, where
 $T_{max}$ is the maximum iteration
number. 

Thus, the overall time complexity of the FRES framework is $O(M)+O(M)+O(M \times T_N)+O(M \times {T_{max}}) \approx O(M \times T_N)$, due to we set $ T_N  \geq T_{max}$. Considering the online decision for all UEs, the $T_N$ of the multi-task agent and the $T_{max}$ of the LTS are set with very small values. Hence,  the time complexity of the FRES framework is low and the FRES framework is suitable for IRS and UAV-assisted MEC system in the dynamic environment.

\section{Experimental Results}
\subsection{Experimental Settings}
The simulation parameters of the FRES framework are chosen as follows: 
The number of UEs is set to 50, and the number of IRSs is also set to match the number of UEs.
The initial multi-task agent consists of two shared layers with 64 and 128 neurons, respectively. Both the fitting head and classification head are configured with 32 neurons \cite {jiang2020ai}. For progressive scheduler, we adjust each layer by adding or removing up to 16 neurons for each UAV.
For the LTS, we set the taboo list length to 5 with a maximum iteration number capped at 10.
Unless otherwise stated, all other parameters employed are summarized in Table \ref{tab2} \cite {jiang2023large,liu2019some,jiang2020ai}. The simulations are executed in an environment powered by an Intel Xeon CPU, supplemented with 32GB RAM, and a Tesla T4 GPU with 15GB RAM.

\begin{table}
	\centering\makegapedcells
	\caption{Experimental parameters.}
	\label{tab2}
	\begin{tabular}{|l|l|}
		\hline
		$\textbf{Parameters}$  & $\textbf{Settings}$ \\ 
		\hline
	
		Maximum number of UAVs $L $ & 5\\
		\hline
		Altitude of the UAV $ Z^V $ & 30 m\\
		\hline
		Altitude of the IRS $ Z^R $ & 15 m\\
		\hline
		Bandwidth $ B $ & 1 MHz\\
		\hline
		Required number of CPU cycles $ F_{i} $ & $ 0.95-1.05 \times 10^{9} $ cycles/s\\
		\hline
		Transmitting data size $ R_{i} $ & $19-21$ MB\\
		\hline
		Transmitting power of UEs $ P_{ij}^{t} $ & 1 W\\
		\hline

		Hovering power of UAVs $ P_{j}^{h} $ & 1 W \\
		\hline
		Local computation capacity $ F_{i,max}^{L} $ &  $10^{9} $ cycles/s \\
		\hline

		Remote computation capacity (UAV)  $ F_{j,max}^{R} $ &  $3\times10^{10} $ cycles/s\\
		\hline
		
	\end{tabular}
\end{table}

\subsection{Experiments for the Multi-task Agent}
\begin{figure}[htpb]
	\centering
	\includegraphics[width=8cm]{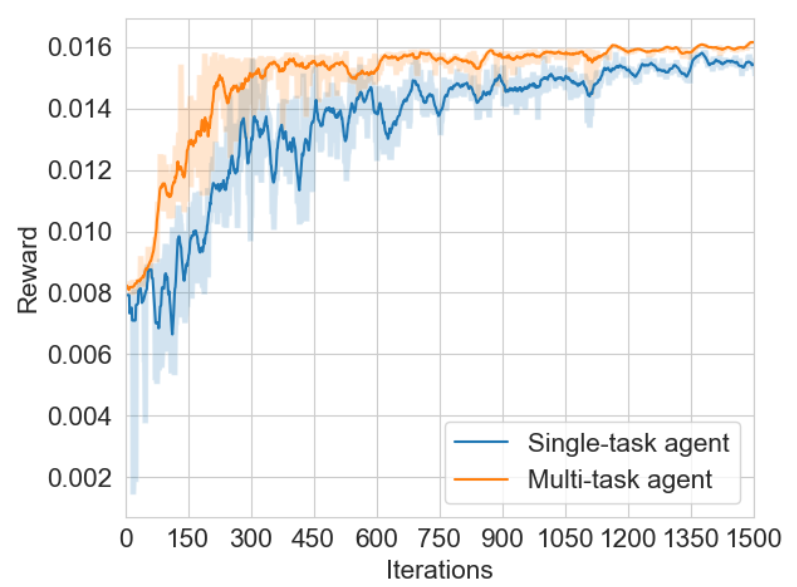}
	\caption{Rewards of single-task agent and multi-task agent.}
	\label{fig:5_1}
\end{figure}

In this section, we verify the effectiveness of the multi-head structure in solving resource scheduling problems by comparing the performance of the single-task agent and the multi-task agent.
For fairness, the number of neurons in single-task agent is set equal to the number of neurons in multi-task agent. 
In the single-task agent, MSE loss is used to optimize all network parameters. The rewards of single-task agent and multi-task agent are shown in Fig. \ref{fig:5_1}. 
Obviously, both single-task and multi-task agents can obtain good rewards and ultimately achieve convergence. However, the multi-task agent is able to obtain higher reward than the single-task agent. This is because task offloading and resource allocation are two different tasks, the single-task agent uses only MSE loss to optimize all network parameters, whereas the multi-task agent uses two different losses to optimize different task heads.

\subsection{Experiments for the Progressive Scheduler}
\begin{figure}[htpb]
	\centering
	\includegraphics[width=8cm]{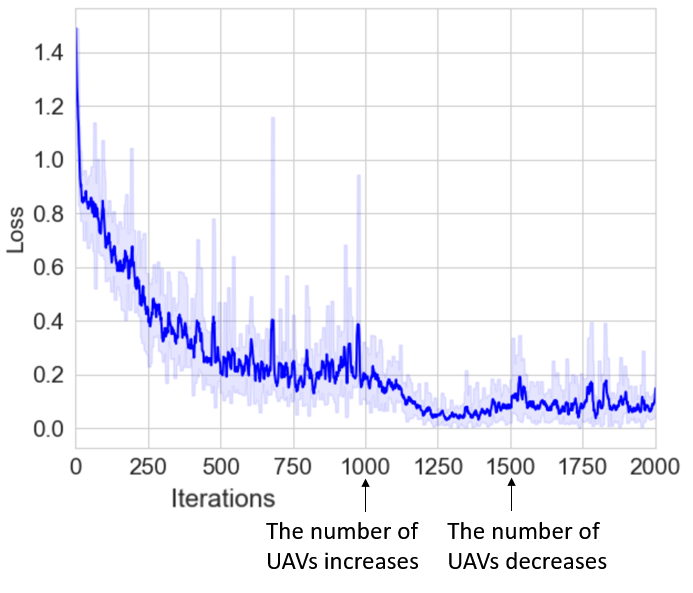}
	\caption{Loss curve for different number of UAVs.}
	\label{fig:5_2}
\end{figure}

\begin{figure}[htpb]
	\centering
	\includegraphics[width=8cm]{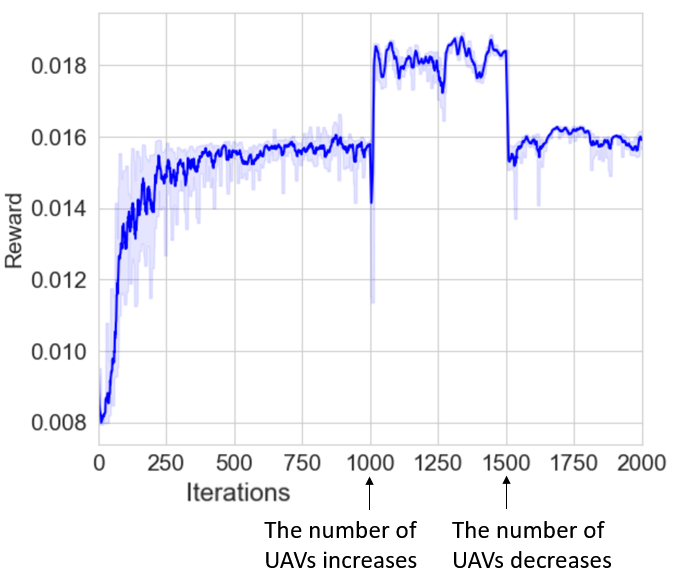}
	\caption{Reward curve for different number of UAVs.}
	\label{fig:5_3}
\end{figure}

In this section, we simulate a scenario where the number of UAVs is varying, and compare the loss and reward before and after the number of UAVs is changed. This simulation is applied to verify the validity of the progressive scheduler.
In the first 1000 iterations, only three UAVs are provided to UEs for mobile computing, in the next 500 iterations, the number of UAVs increases to four, and in the last 500 iterations, the number of UAVs is back to three again. The loss curve and the reward curve are plotted in Fig. \ref{fig:5_2} and Fig. \ref{fig:5_3}. 

In Fig. \ref{fig:5_2}, it can be seen that in the first 1000 iterations, the loss of the agent continues to decrease to convergence. As the number of UAVs increases, the adding rule is carried out, and the knowledge of the new structure of the agent can be used in the new scenario with four UAVs directly. When the number of UAVs returns to three again, the removing  rule is executed and the extra structure of the agent is closed. The agent do not need to retrain again and the loss curve do not alter significantly throughout the process.

Similar to Fig. \ref{fig:5_2}, Fig. \ref{fig:5_3} shows the reward of the agent during the number of UAVs changes. In the first 1000 iterations, the reward continues to increase and then achieves convergence. When the number of UAVs increases, the adding rule is carried out and the overall reward is changed. When the number of UAVs returns to three, the removing rule is executed and the extra structure of the agent is closed, so the reward is back to the original level.

\subsection{Experiments for the LTS}
In this section, we evaluate the performance of the LTS with three different single point search methods to verify the exploration of the LTS. The three heuristic search methods are TS \cite{RN26}, simulated annealing (SA)\cite{sa} and adaptive SA (ASA) \cite{biswas2021mcdm}.
In the experiment, the iteration number of LTS and TS is set to 90, the length of taboo list is set to 5, the search neighborhood size is set to 20; The annealing cooling rate of SA and ASA is set to 0.95, the initial annealing temperature $ T_{max}=100 $ and cut-off annealing temperature $ T_{min}=1 $. 

\begin{figure}[htpb]
	\centering
	\includegraphics[width=9cm]{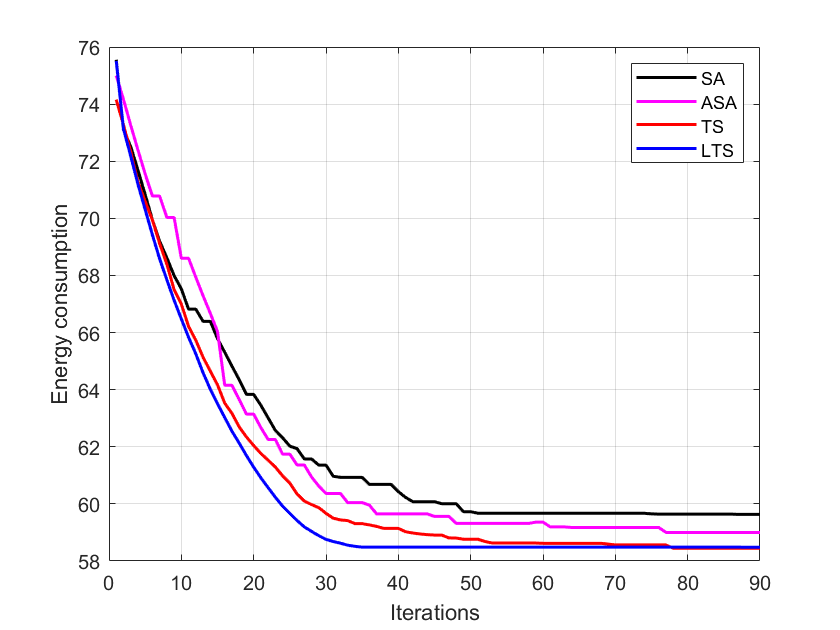}
	\caption{Energy consumption of different exploration methods.}
	\label{fig:5_4}
\end{figure}

Fig. \ref{fig:5_4} presents the performance of different action refinements. We can see that LTS and TS obtain lower energy consumption than ASA and SA, while LTS converges faster than TS. This is due to the fact that LTS employs the light move operator with the highest channel gain to accelerate the search.

\subsection{Experiments for the FRES Framework}
In this section, we evaluate the performance of the whole FRES framework. We first compare the FRES framework with four 
DRL algorithms: DDPG \cite{qiu2019deep}, SAC \cite{haarnoja2018soft}, LyDROO \cite{bi2021lyapunov} and ARE \cite {jiang2020ai}. TABLE \ref{tab:table2} characterizes the training time of initial task (only one UAV at the beginning), new task (adding a new UAV), old task (removing a UAV), and the average and standard deviation (STD) of the energy consumption in all DRLs. It can be seen that the FRES framework achieves the least training time of the old task and the lowest average energy consumption. The superiority of the FRES framework can be explained as follows: (1) The progressive scheduler can reuse the stored policy knowledge when the number of UAVs is changing without retraining the agent again. Hence, the training time of the old task in the FRES framework is the least.
(2) LTS refines the action and boosts exploration capabilities, enabling it to escape local extrema during the search process. As a result, the FRES framework achieves the lowest average energy consumption.

\begin{figure*}[ht]
	\centering
	\includegraphics[width=18cm]{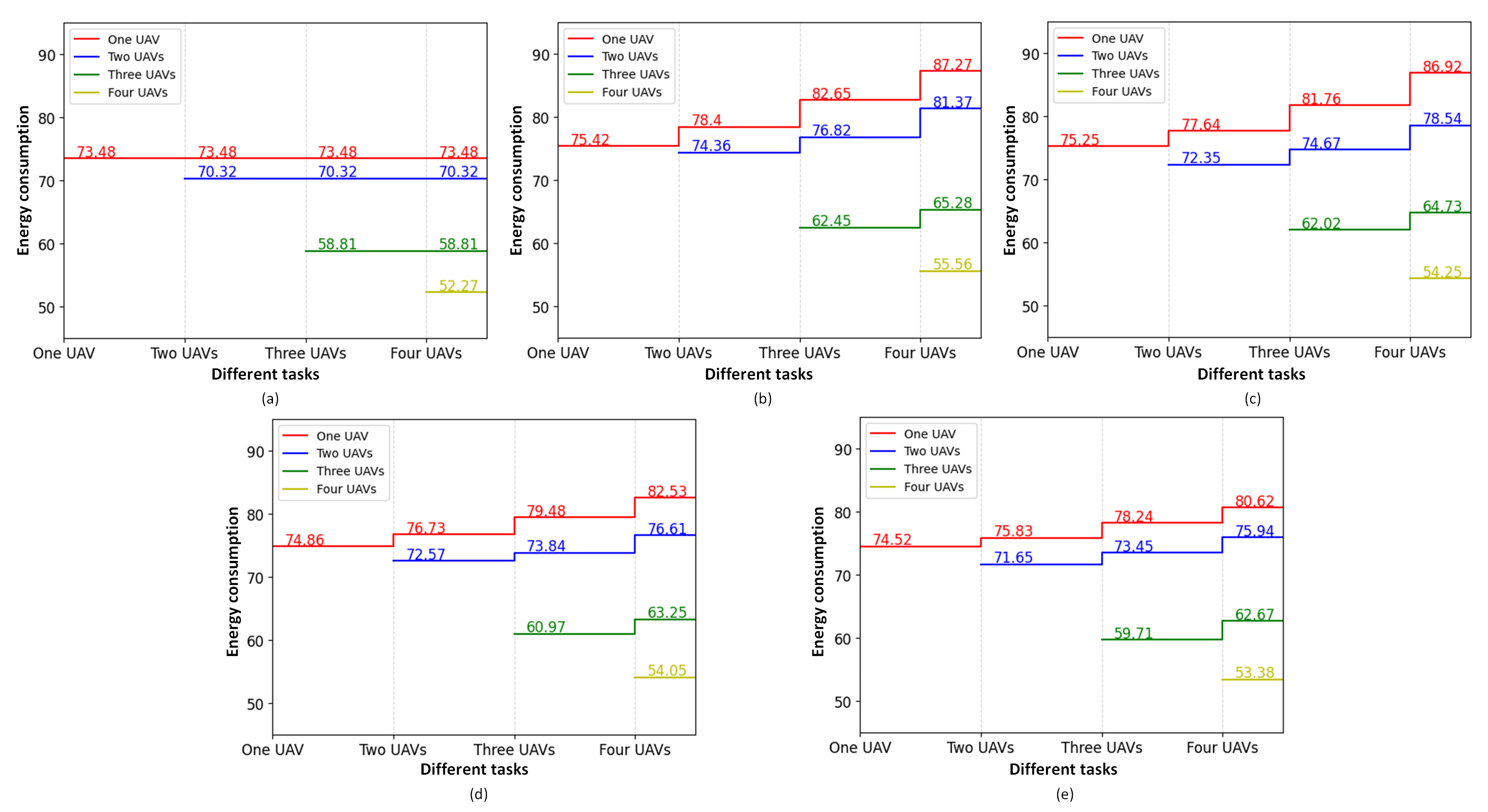}
	\caption{Multi-task performances of different DRLs. (a)FRES. (b)DDPG. (c)SAC. (d)LyDROO. (e)ARE.}
	\label{fig:6}
\end{figure*}

Then, we use the multi-task map proposed by Google to illustrate the performance variations of old tasks when the new task is added. Fig. \ref{fig:6} summarizes the multi-task performances of different DRLs, in which we add three UAVs as new tasks one by one to the DRLs. Different colored lines describe the change in energy consumption of the old tasks when a new task is added. As seen in Fig. \ref{fig:6}, while training for a new task (adding a new UAV), the energy consumption on older tasks increases continuously in the cases of DDPG, SAC, LyDROO and ARE, which means the catastrophic forgetting has already happened, and the knowledge of old task has been forgotten. Our FRES is superior to the other DRLs, and the energy consumption on all tasks remains fixed.

\begin{table}[]
	\centering\makegapedcells
	\caption{Performance of different DRLs for training time and energy consumption.}	
	\label{tab:table2}
	\begin{tabular}{|p{25pt}|p{30pt}p{30pt}p{30pt}<{\centering}|p{30pt}p{30pt}<{\centering}|}
		\hline
		\multirow{2}{*}{Method} & \multicolumn{3}{c|}{Training time}                            & \multicolumn{2}{c|}{Energy consumption}    \\ \cline{2-6} 
		& \multicolumn{1}{c|}{Initial task} & \multicolumn{1}{c|}{New task} & Old task & \multicolumn{1}{c|}{Average} & STD \\ \hline
		FRES	& \multicolumn{1}{c|}{221.14} & \multicolumn{1}{c|}{114.32} & 0.43 & \multicolumn{1}{c|}{56.37} & 1.47 \\ \hline
		DDPG	& \multicolumn{1}{c|}{254.36} & \multicolumn{1}{c|}{183.42} & 128.76 & \multicolumn{1}{c|}{61.55} & 2.65 \\ \hline
		SAC	& \multicolumn{1}{c|}{242.65} & \multicolumn{1}{c|}{178.47} & 118.28 & \multicolumn{1}{c|}{60.09} & 2.37 \\ \hline
		LyDROO	& \multicolumn{1}{c|}{245.33} & \multicolumn{1}{c|}{157.65} & 66.72 & \multicolumn{1}{c|}{58.81} & 2.04 \\ \hline
		ARE	& \multicolumn{1}{c|}{262.47} & \multicolumn{1}{c|}{136.93} & 52.23 & \multicolumn{1}{c|}{58.26} &  1.88\\ \hline
	\end{tabular}
\end{table}

Finally, we select five offloading schemes as benchmarks to evaluate the performance of our offloading approach. The chosen benchmarks are described as follows:

\begin{itemize}	
	\item Random offloading (Random)  denotes that the offloading decision of each UE is randomly determined.
	\item Local executed (Local) denotes that all UEs execute tasks locally.
	\item  Offloading nearby (Remote) denotes that each UE decides to offload the task to the closest UAV.
    \item TS denotes that taboo search is applied to search the best offloading decision for all UEs.
    \item 
    Lyapunov-guided deep reinforcement learning online offloading  (LyDROO) is a celebrated DRL offloading scheme for the MEC system \cite{bi2021lyapunov}. 
\end{itemize}

\begin{figure}[htpb]
	\centering
	\includegraphics[width=9cm]{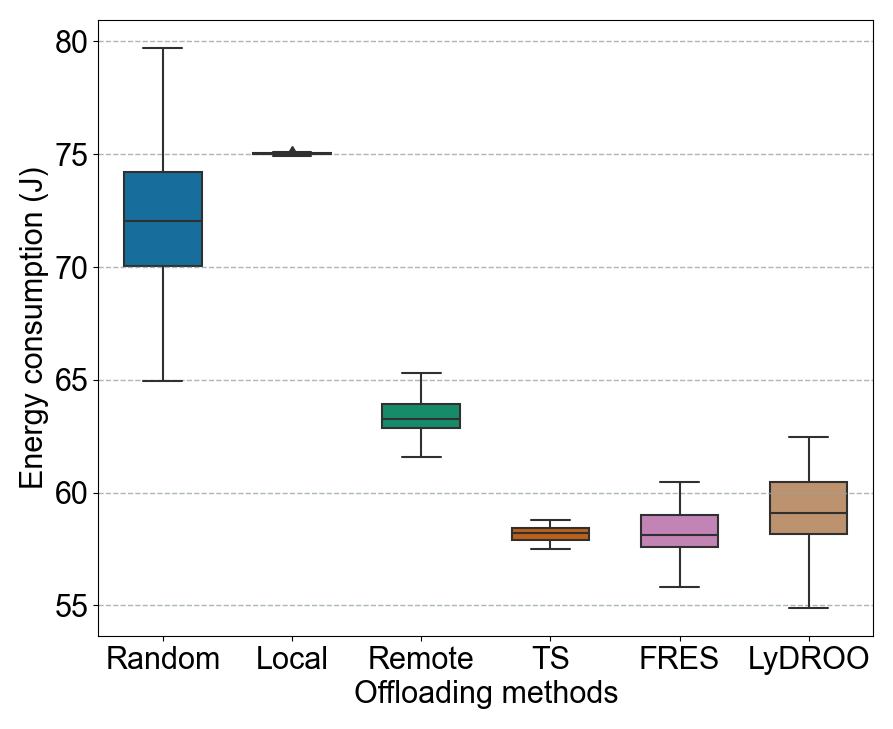}
	\caption{Energy consumption of different task offloading methods.}
	\label{fig:5_5}
\end{figure}

The average and STD of the energy consumption for these five offloading methods are shown in Fig. \ref{fig:5_5}, it can be seen that the energy consumption of the proposed FRES framework is much lower than the energy consumptions of Local, Random and Remote. Meanwhile, the energy consumption of the proposed FRES framework is closed to the TS. This is because the FRES method can update the offloading policy from the high-quality offloading solutions generated by LTS. 
Furthermore, the FRES method is capable of constructing a multi-task agent to generate offloading decisions and resource allocation, which can make faster high-quality decisions than traditional heuristic search methods.

\section{Conclusion}
In this paper, we propose a deep progressive reinforcement learning based FRES framework to minimize the energy consumption of all the UEs by optimizing offloading decision and resource allocation in IRS and UAV-assisted MEC system. The FRES framework includes three important parts: 1) the multi-task agent is presented to solve the offloading decision task and the resource allocation task at the same time; 2) the progressive scheduler is used to achieve quick reaction for the dynamic environment with the changing number of UAVs; 3) the LTS is applied to enhance the exploration of the DRL. The simulation results show that the FRES framework can achieve the best performance in IRS and UAV-assisted MEC system with varying number of UAVs.

Despite the contributions of the proposed method, there is also a downside that can serve as a basis for future research. When the number of UAVs increases, the   number of parameters in the multi-task agent is also growing. This growth can be addressed by pruning, quantization and compression for the network structure of the agent during the learning process.


%

%




\bibliographystyle{ieeetran}
\bibliography{bare_jrnl_wang}

\section*{Biographies}
\begin{IEEEbiography}[{\includegraphics[width=1in,height=1.25in,keepaspectratio]{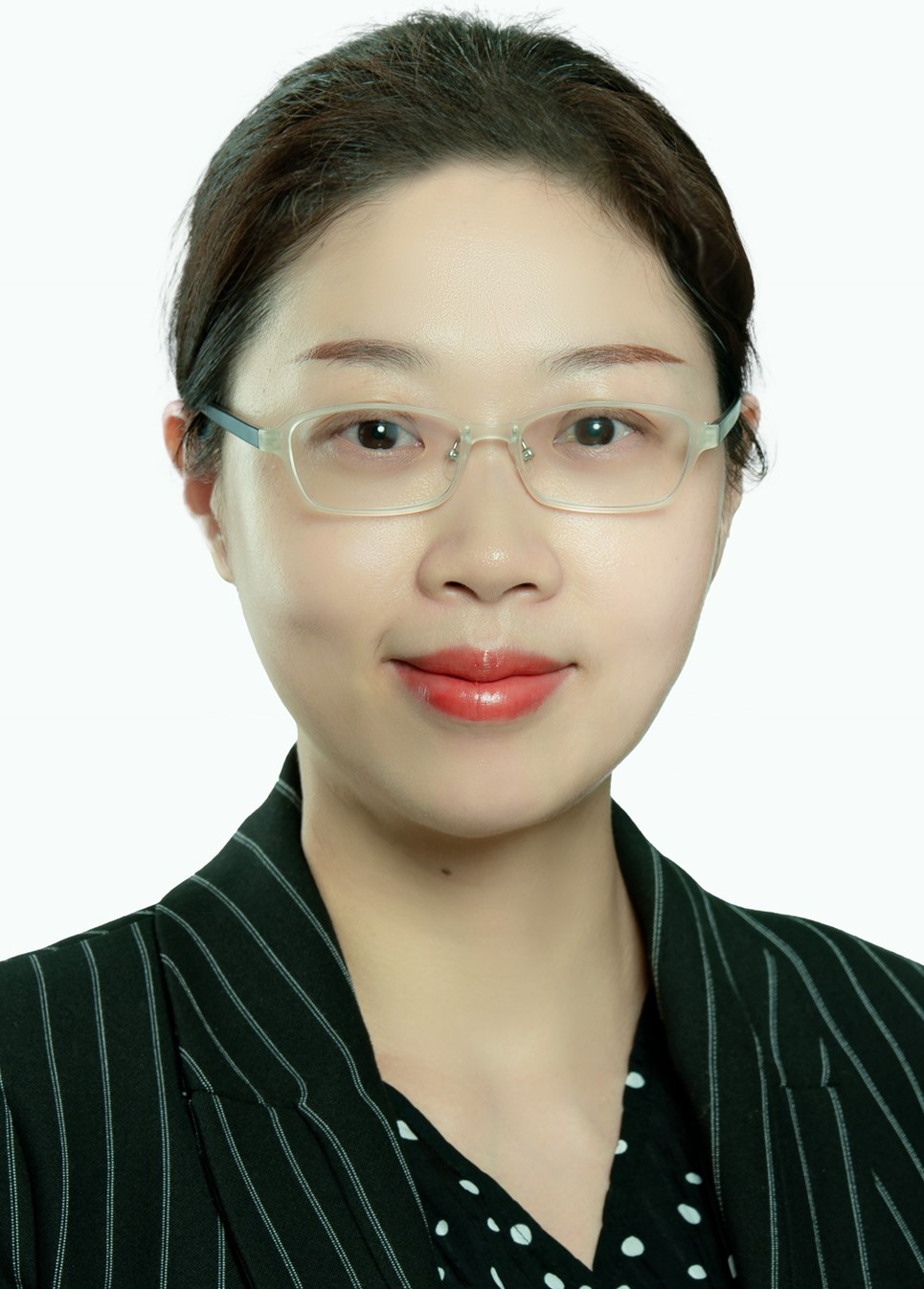}}]{Li Dong} received the B.S. and M.S. degrees in School of Physics and Electronics from Hunan Normal University, China, in 2004 and 2007, respectively. She received her Ph.D. degree in School of Geosciences and Info-physics from the Central South University, China, in 2018. She is currently an associate professor at Hunan University of Technology and Business, China. Her research interests include machine learning, Internet of Things, and mobile edge computing.
\end{IEEEbiography}
\vspace{-10 mm}
\begin{IEEEbiography}[{\includegraphics[width=1in,height=1.25in,keepaspectratio]{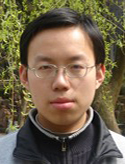}}]{Feibo Jiang} received his B.S. and M.S. degrees in School of Physics and Electronics from Hunan Normal University, China, in 2004 and 2007, respectively. He received his Ph.D. degree in School of Geosciences and Info-physics from the Central South University, China, in 2014. He is currently an associate professor at the Hunan Provincial Key Laboratory of Intelligent Computing and Language Information Processing, Hunan Normal University, China. His research interests include federated learning, semantic communication, Internet of Things, and mobile edge computing.
\end{IEEEbiography}
\vspace{-10 mm}
\begin{IEEEbiography}[{\includegraphics[width=1in,height=1.25in,keepaspectratio]{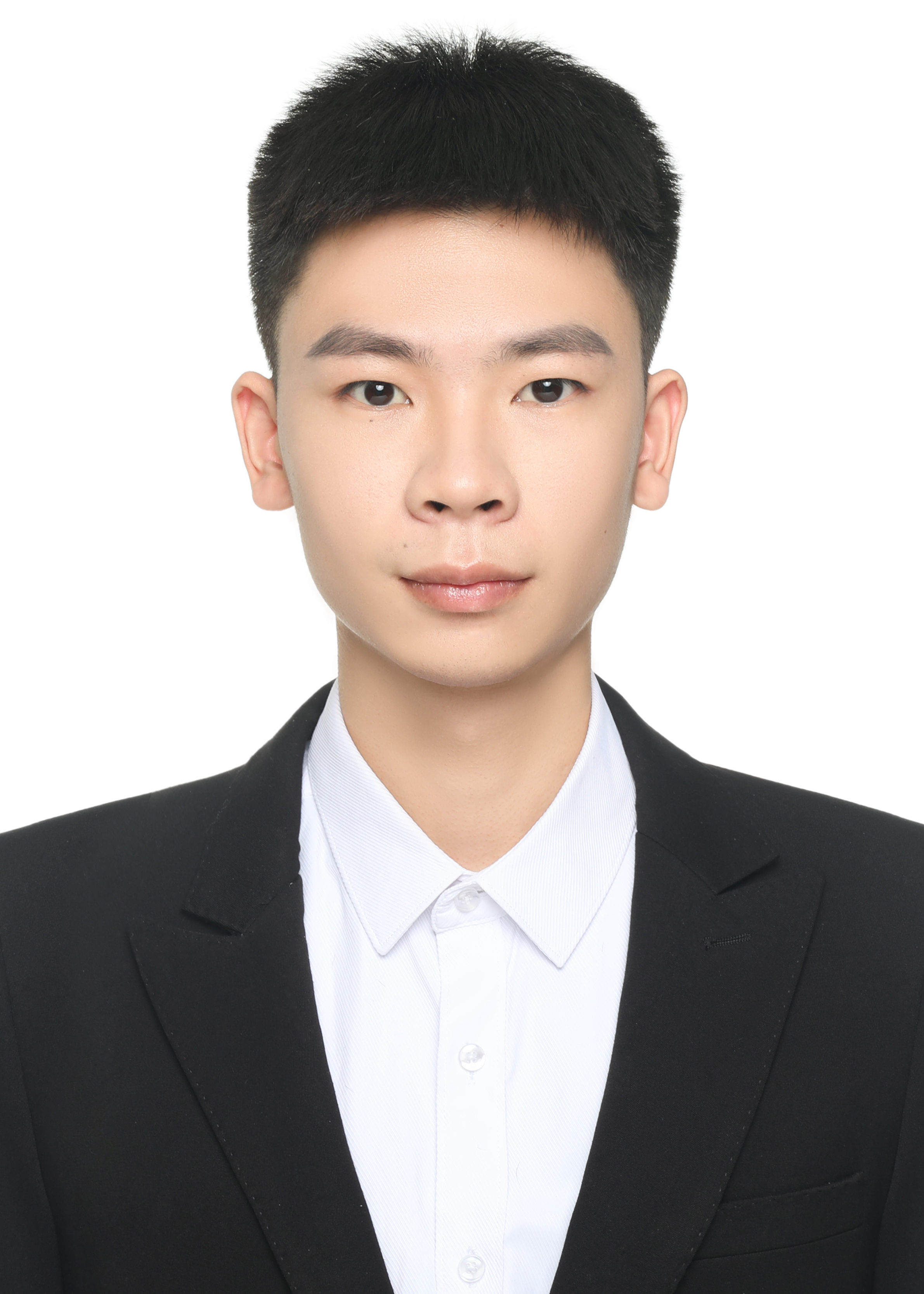}}]{Minjie Wang} received the M.S. degree from Hunan Normal University, Changsha, China, in 2022. His main research interests include pattern recognition and intelligence system.
\end{IEEEbiography}
\vspace{-10 mm}
\begin{IEEEbiography}[{\includegraphics[width=1in,height=1.25in,keepaspectratio]{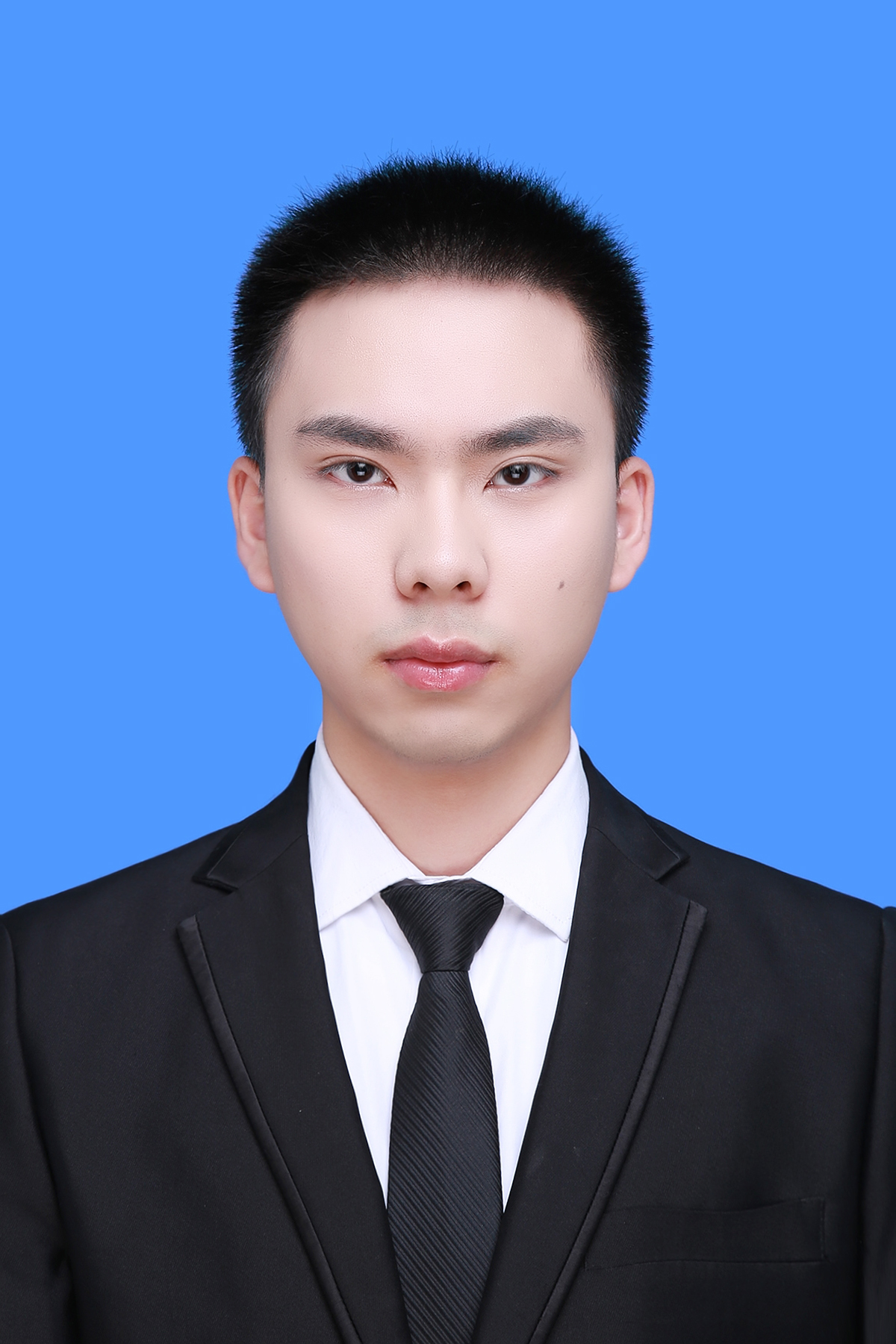}}]{Yubo Peng} received the B.S. degree from Hunan Normal University, Changsha, China, in 2019, where he is currently pursuing the master’s degree with the College of Information Science and Engineering. His main research interests include federated learning and semantic communication.
\end{IEEEbiography}
\vspace{-10 mm}
\begin{IEEEbiography}[{\includegraphics[width=1in,height=1.25in,keepaspectratio]{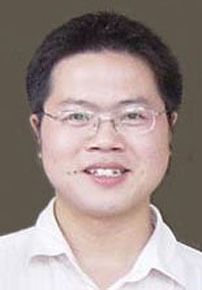}}]{Xiaolong Li} received the B.E. degree from the Harbin Institute of Technology, Harbin, China in 2003 and the Ph.D. degree from the Hunan University, Changsha in 2008. From September 2008 to December 2016, he was with the School of Computer Science and Engineering, Guilin University of Electronics and Technology, Guilin, China, where he was a Full Professor. Since January 2017, he has been with the School of Computer Science, Hunan University of Technology and Business, Changsha, China, as a Full Professor and Dean of School of Computer Science. In 2013, he won the best paper award at the ChinaCom conference. His research interests include industrial IoT, edge/fog Computing, and aritificial intelligence.
\end{IEEEbiography}

\newpage
\end{document}